\documentclass{article}

\usepackage[final]{neurips_2025}
\makeatletter
\newcommand{\@trackname}{}
\makeatother
\usepackage{multirow}
\usepackage[table,xcdraw]{xcolor}
\usepackage{colortbl}
\usepackage{url}
\usepackage[T1]{fontenc}
\usepackage{amsmath}
\usepackage{natbib} 
\usepackage{enumitem}
\usepackage{algorithm}
\usepackage{algorithmicx}
\usepackage{algpseudocode}
\usepackage{algpseudocode}
\usepackage{amsmath, amssymb}
\usepackage{svg}
\usepackage{booktabs}
\usepackage{placeins}
\usepackage{tcolorbox}
\usepackage[utf8]{inputenc}
\usepackage{enumitem} 
\usepackage{booktabs}
\usepackage{subcaption} 

\usepackage[utf8]{inputenc} 
\usepackage[T1]{fontenc}    
\usepackage{hyperref}       
\usepackage{url}            
\usepackage{booktabs}       
\usepackage{amsfonts}       
\usepackage{nicefrac}       
\usepackage{microtype}      
\usepackage{xcolor}         

\title{Go-UT-Bench: A Fine-Tuning Dataset for LLM-Based Unit Test Generation in Go}

\author{%
Yashaswi Piplani \thanks{Equal contribution.} \\
  Nutanix\\
  \texttt{yashaswi.piplani@nutanix.com} \\
  \And
  Hritik Raj \footnotemark[1] \\
  Nutanix\\
  \texttt{hritik.raj@nutanix.com} 
  \And
  Rajat Ghosh \\
  Nutanix\\
  \texttt{rajat.ghosh@nutanix.com} 
  \And
  Vaishnavi Bhargava \\
  Nutanix\\
  \texttt{vaishnavi.bhargava@nutanix.com}
  \And
  Debojyoti Dutta \\
  Nutanix\\
  \texttt{debojyoti.dutta@nutanix.com}
}

\begin{document}

\maketitle

\begin{abstract}

Training data imbalance poses a major challenge for code LLMs: most available code heavily overrepresents raw open-source code while underrepresenting broader software engineering tasks, especially in low-resource languages like Golang. As a result, models excel at code autocompletion but struggle with real-world developer workflows such as unit test generation. To address this gap, we introduce Go-UT-Bench, a benchmark dataset of 5,264 \{code, unit test\} pairs drawn from 10 permissively licensed Golang repositories spanning diverse domains including distributed systems, cloud infrastructure, blockchain, and API development. We evaluate its effectiveness as a fine-tuning dataset across two LLM families—Mixture-of-Experts (e.g., DeepSeek-Coder-V2-Lite-Instruct) and dense decoders (e.g., Llama-3.2-3B-Instruct). Our results show that fine-tuned models outperform their base counterparts on more than 75\% of benchmark tasks. We release Go-UT-Bench with commit hashes for full reproducibility and highlight challenges in dataset design and fine-tuning trade-offs.

\end{abstract}

\section{Introduction}
\label{lab:intro}

Large language models (LLMs) are typically pretrained as next-token predictors and later adapted to instruction-following tasks through post-training~\citep{nijkamp2023codegenopenlargelanguage}. A common post-training approach is supervised fine-tuning (SFT), supported by libraries such as Hugging Face TRL, PEFT, and Meta’s torchtune. While the methods and tooling for supervised fine-tuning (SFT) are well established, the primary bottleneck lies in obtaining high-quality, task-specific datasets. This challenge is amplified in domain-specific tasks where data is scarce, limited in diversity, or lacks permissive licensing. Recent work highlights that the lack of high-quality, task-specific data becomes a bottleneck for SFT, limiting the potential of LLMs to excel in specialized applications~~\citep{zhang2023instructiontuning,li2024apiguided,lin2025domainrag}. In the context of code-focused LLMs, synthesizing robust datasets remains particularly difficult, as even state-of-the-art efforts note the uneven quality and scarcity of domain-specific supervised fine-tuning corpora~~\citep{li2024apiguided}. For example, constructing a fine-tuning dataset for Go unit test generation is particularly difficult due to the scarcity of openly available, diverse, and permissively licensed \{code, unit test\} pairs and the additional requirement of reproducibility for transparent research. The current closed-source data policy of leading GenAI organizations presents a formidable challenge to the future of open LLM research. Although there have been few initiatives~~\citep{lambert2025tulu3pushingfrontiers, li2023starcodersourceyou,lozhkov2024starcoder2stackv2}, but that is far from sufficient.

Go (often referred to as Golang), a statically typed, compiled language, is widely used in industry for building scalable and efficient software systems. Its concurrency model, robust error handling, and simplicity make it a popular choice, but unique features such as Goroutines, channels, and interfaces add complexity. Writing and maintaining unit tests in Go is therefore challenging, creating a need for automation that can be supported by fine-tuned LLMs. However, despite Go’s industrial importance, there is a lack of reproducible fine-tuning datasets for unit test generation.

To address this gap, we introduce Go-UT-Bench, a large-scale, extensible fine-tuning dataset of \{code, unit test\} pairs curated from diverse Go repositories. We evaluate its effectiveness by showing that LLMs fine-tuned on Go-UT-Bench generate higher-quality unit tests than their base counterparts. Our key contributions are as follows:

\begin{itemize}
    \item An extensible and reproducible fine-tuning dataset for unit test generation in Go, with sufficient scale and domain diversity to support adaptation of LLMs.
    \item A reproducible data curation pipeline for extracting {code, unit test} pairs from permissively licensed open-source repositories.
\end{itemize}

\section{Go-UT-Bench as a Golang Unit Test Benchmark Dataset}

\label{lab:data}
Go-UT-Bench is a fine-tuning dataset for the task of unit test generation in Go, formalized as the mapping $\mathcal{C} \rightarrow \mathcal{T}$, where $\mathcal{C}$ denotes a Go source file and $\mathcal{T}$ its corresponding unit test. Figure \ref{fig:code-unit-test-example} in Appendix shows an example of \{code, unit test\} pairs. The dataset contains 5{,}264 \{code, unit test\} pairs, designed to provide sufficient scale for model adaptation. Fine-tuning typically benefits from datasets in the range of 1{,}000--10{,}000 examples~~\citep{spaceo2024llm, predibase2024finetune, zhou2023lima, barnett2024openai, huggingface2022finetune}, placing Go-UT-Bench within the recommended scale for effective instruction- or task-specific adaptation. In addition to the code file $\mathcal{C}$ and test file $\mathcal{T}$, Go-UT-Bench provides metadata fields to facilitate reproducibility and provenance tracking:

\begin{itemize}[leftmargin=*]
\item \textbf{SHA256}: A unique identifier for each entry, computed from the code file.
\item \textbf{Repository}: The GitHub repository in the format \texttt{organization/repository} (e.g., \texttt{kubernetes/kubernetes}).
\item \textbf{File Name}: The base name of the file containing the code or test (e.g., \texttt{sample1}).
\item \textbf{File Path in Repository}: The relative path of the file within the repository (e.g., \texttt{pkg/controlplane/controller/leaderelection/election.Go}).
\item \textbf{Code}: The full source code content of the Go file.
\item \textbf{Code Commit Hash}: The Git commit hash corresponding to the version of the code file.
\item \textbf{File Path for Unit Test}: The relative path to the associated unit test file (e.g., \texttt{pkg/controlplane/controller/leaderelection/election\_test.Go}).
\item \textbf{Unit Test (Ground Truth)}: The full content of the unit test file.
\item \textbf{Unit Test Commit Hash}: The Git commit hash corresponding to the version of the unit test file.
\end{itemize}

A key contribution of Go-UT-Bench is that it is constructed from real-world, industrial-scale Go projects rather than synthetic or toy examples. As summarized in Table~\ref{tab:Go_ut_bench_domain}, the dataset spans widely adopted open-source repositories across diverse domains. For instance, it includes large-scale infrastructure systems such as Kubernetes (1,935 tests), PingCap TiDB (878 tests), and Golang itself (761 tests), alongside major cloud and container frameworks like Terraform (482 tests) and Moby (412 tests). It further incorporates high-impact domain-specific projects, including Ethereum for blockchain (285 tests), HuGo for website creation (226 tests), Prometheus for metrics monitoring (185 tests), KServe for model serving (73 tests), and Gin for API development (27 tests). These repositories are actively maintained and deployed in production environments, reflecting authentic software engineering practices such as concurrency management, modularity, and large-scale test organization. By capturing both dominant infrastructure projects and long-tail distribution, Go-UT-Bench provides a fine-tuning dataset for unit test generation in Go that explicitly represents real-world software engineering challenges.
        
\renewcommand{\arraystretch}{1.5}

\begin{table}[ht]
\centering
\caption{Diversity of Go-UT-Bench: Varied repositories, their test counts in Go-UT-Bench, and corresponding domains.}
\label{tab:Go_ut_bench_domain}
\begin{tabular}{lrl}
\toprule
\textbf{Repository} & \textbf{Count in Go-UT-Bench} & \textbf{Domain} \\ 
\midrule
Kubernetes & 1935 & Container Orchestration \\ 
PingCap TiDB & 878 & Distributed SQL Database \\ 
Golang & 761 & Concurrent Programming \\ 
Hashicorp Terraform & 482 & Cloud Resource Management \\ 
Moby & 412 & Container Framework Toolkit \\ 
Ethereum & 285 & Blockchain \\ 
HuGo & 226 & Website Creation \\ 
Prometheus & 185 & Metrics Monitoring \\ 
Kserve & 73 & Model Serving \\ 
Gin-Gonic Gin & 27 & API Development Framework \\ 
\bottomrule
\end{tabular}
\end{table}

\begin{figure}[h!]
    \centering    \fbox{\includegraphics[width=0.6\textwidth]{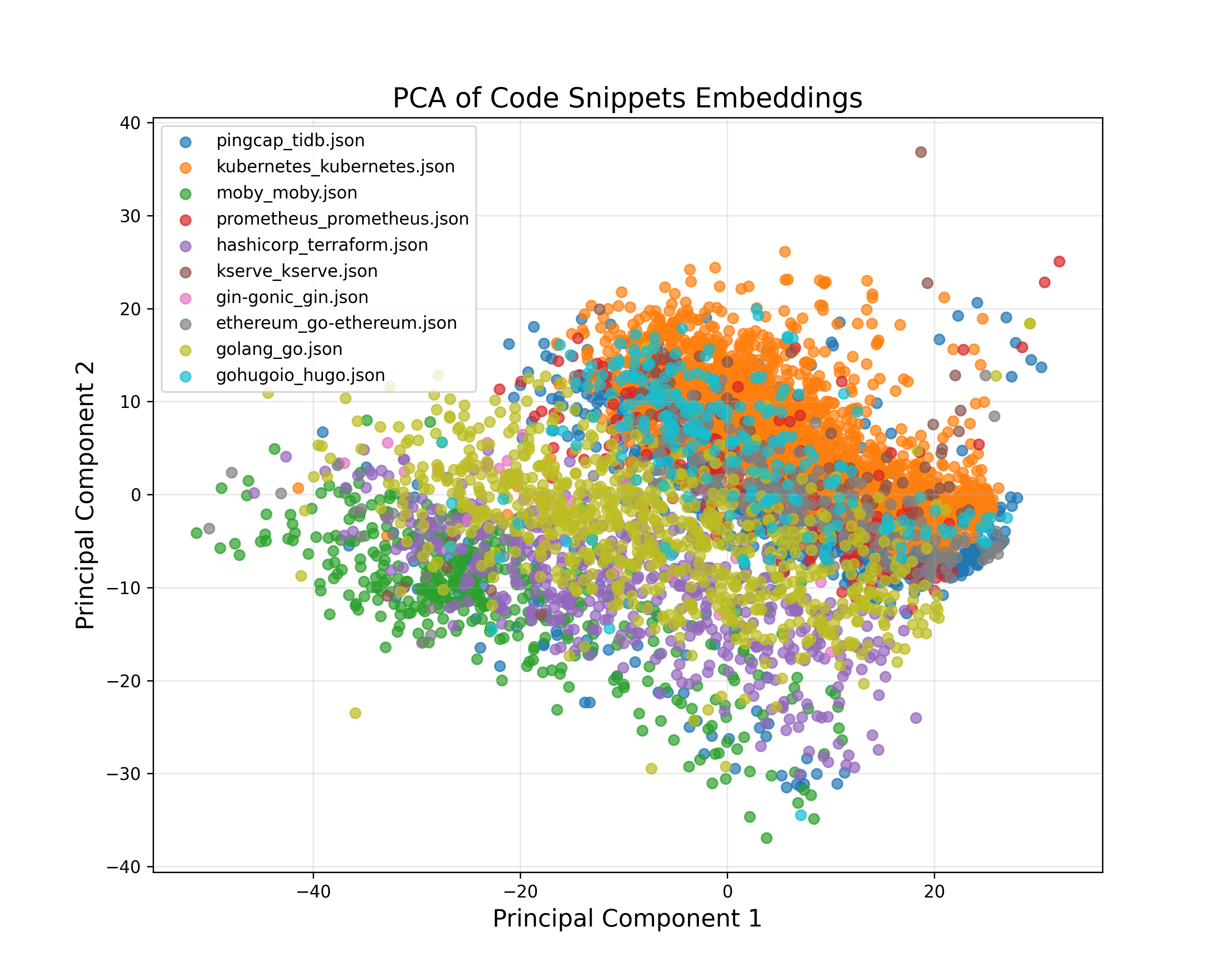}}
    \caption{PCA based analysis of Go-UT-Bench revealing the internal structure of the dataset.  }
    \label{fig:pca_code_scatter_plot}
\end{figure}

Go-UT-Bench is a heterogeneous dataset constructed from ten open-source repositories with imbalanced contributions. In contrast to text data, code data are inherently anisotropic, characterized by recurring structural motifs (e.g., Goroutines) and frequent syntactic tokens (e.g., \textit{fmt}), which are often connected with multiple logical hops. To quantify this anisotropy and repository-level skewness, we analyzed the embedding space of the \texttt{code} field in Go-UT-Bench. Figure~\ref{fig:pca_code_scatter_plot} illustrates the top two principal components of this embedding space, revealing both heterogeneity and distributional imbalance across repositories. Snippets from some repositories (e.g., \texttt{kubernetes/kubernetes}, \texttt{prometheus/prometheus}) form coherent clusters, indicating possible semantic similarities. At the same time, partial overlaps are visible across repositories, reflecting shared Go idioms and common libraries. This distribution highlights both the domain diversity and structural coherence of the dataset—two properties that make Go-UT-Bench particularly well-suited for fine-tuning models on the task of unit test generation.

\begin{figure}[h!]
    \centering    \fbox{\includegraphics[width=0.8\textwidth]{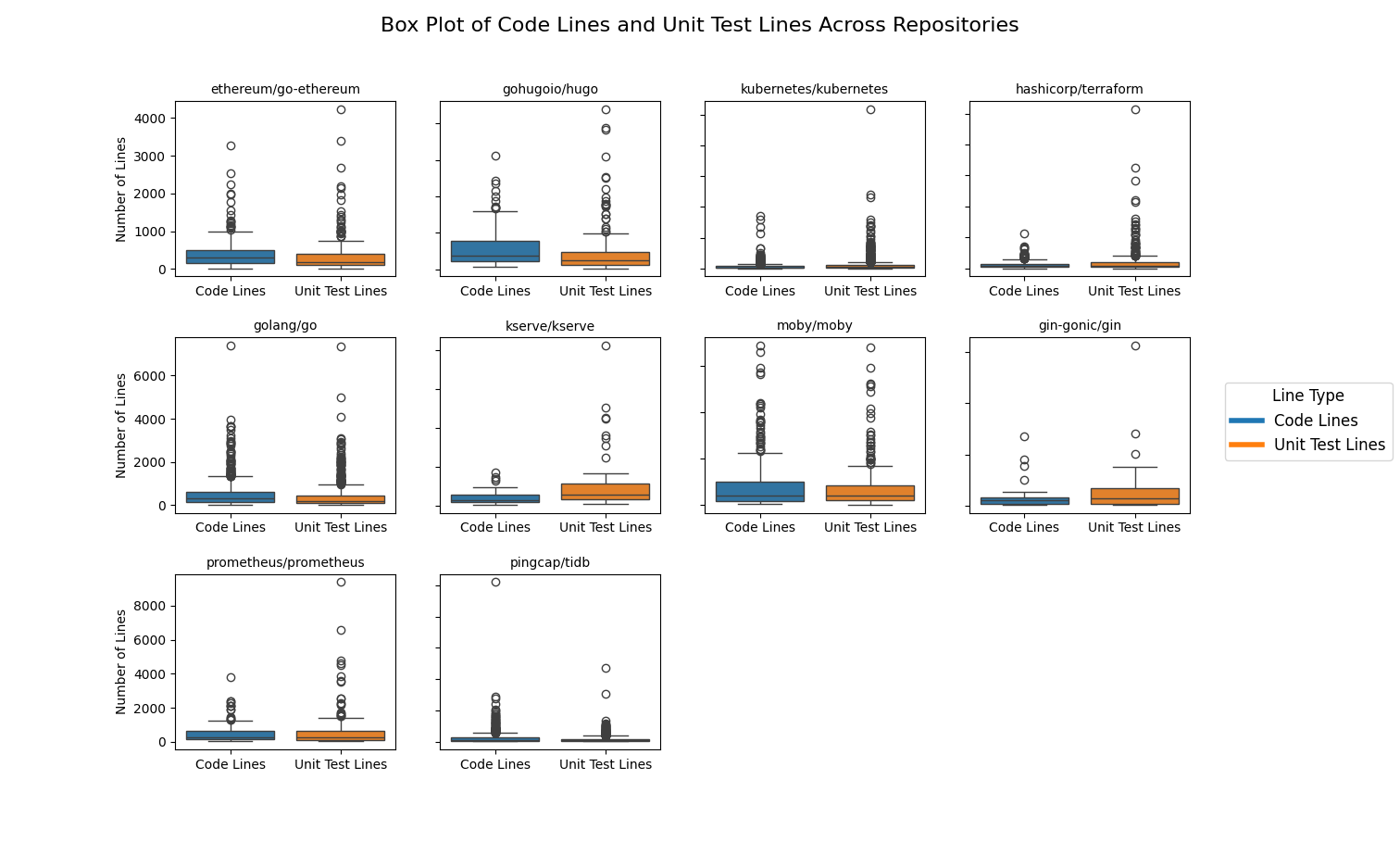}}
    \caption{The diversity in \{code, unit test\} pairs in terms of line lengths across 10 different opensource repositories in Go-UT-Bench.  }
    \label{fig:relative_loc_diversity}
\end{figure}

 On a deeper analysis of the dataset, we found that one of the main features revealed with the PCA is the number of lines in the code. For example, Prometheus and HuGo have overlapping areas with both repositories have most of their files in the range 500-1500 code lines. Moby on the other hand have most of its files in the range 20-200 code lines and hence forms a distinct cluster away from Prometheus and HuGo. Golang has files ranging from 20-1500 and even beyond, and hence it overlaps across all regions. We analyzed the distribution of lengths for \{code, unit test\} pairs in Go-UT-Bench, categorized by different repositories, as shown in Figure \ref{fig:relative_loc_diversity}. The results indicate that all repositories exhibit average line lengths exceeding 100, with significant variance and outliers, reflecting the diversity typically observed in real-world codebases.

\subsection{Scalable Data Curation Pipeline}

To construct Go-UT-Bench, we designed a scalable data curation pipeline capable of handling diverse Go codebases. The pipeline follows a two-stage workflow:

\begin{itemize}[leftmargin=*]
    \item \textbf{File Extraction and Grouping:} We begin by scanning each repository recursively to identify all Go source files (``.go'') and unit test files (``\_test.go''). Files are grouped by their base names (excluding extensions), enabling a direct mapping between source files and their associated test files. This ensures comprehensive coverage of all relevant code–test pairs.
    
    \item \textbf{Mapping and Metadata Augmentation:} Each source file is systematically paired with its corresponding test file(s). The resulting \{code, test\} pairs are enriched with metadata—including repository name, relative file path, commit hash, and a SHA256-based unique identifier. These fields guarantee reproducibility and enable precise version tracking across evolving codebases.
\end{itemize}

\section{Experiment Design}
\label{lab:experiment}

 We assessed the effectiveness of Go-UT-Bench as a fine-tuning dataset by evaluating whether a fine-tuned LLM ~\citep{hu2021loralowrankadaptationlarge} can generate higher-quality unit tests compared to its base model. To assess the performance of large language models (LLMs) in unit test generation, we adopt the LLM-as-a-Judge paradigm ~\citep{zheng2023judging}, using GPT-4o-mini as the oracle model. This approach addresses the limitations of conventional NLP metrics such as BLEU and ROUGE, which fail to effectively capture the semantic similarity necessary for evaluating generated code ~\citep{chen2021evaluating}.

\begin{figure}[ht!]
    \centering    
\fbox{\includegraphics[width=\textwidth]{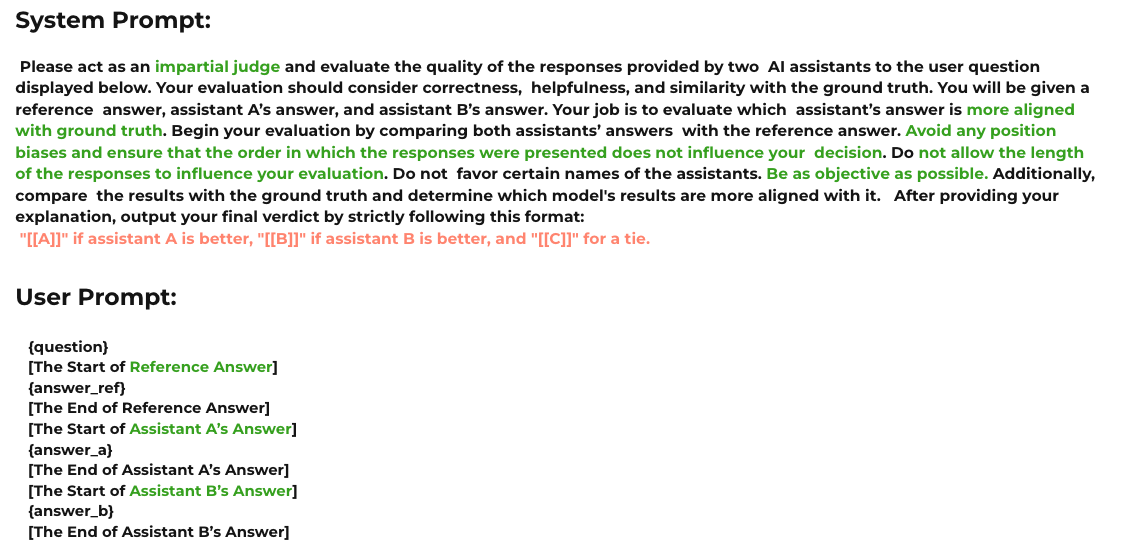}}
    \caption{Prompt for pairwise evaluation of two LLM generated responses (Assistant A and Assistant B) w.r.t. the ground truth.}
    \label{fig:evalprompt}
\end{figure}

Our evaluation framework, formally defined in Algorithm \ref{alg:oracle_eval_tie}, is an oracle-based. pairwise evaluation procedure. It compares responses from two models, Model-A and Model-B, on a dataset $\mathcal{D}$. For each input $x_i$ with ground truth $g_i$, the oracle $J$ computes alignment scores for both responses. 
If one score is higher, that model is declared the winner; if the scores are equal, the outcome is recorded as a tie. 
Over the full dataset, the algorithm tracks the number of wins for each model and the number of ties, and reports the corresponding win and tie rates. 
Figure \ref{fig:evalprompt} illustrates the structured prompt template used for the Oracle evaluation. The prompt has been carefully designed to highlight subtle differences between model outputs and their alignment with the ground truth. To mitigate potential biases, such as GPT’s preference for longer responses or positional bias, we refined the prompt to ensure a fair comparison. In our pairwise evaluation approach, the win rate for a model is defined as the percentage of instances in which its response is judged to be more closely aligned with the ground truth than that of the competing model. This methodology is supported by prior studies demonstrating that GPT-based evaluation closely mirrors human judgment while being more cost-effective and time-efficient ~\citep{zheng2023judgingllmasajudgemtbenchchatbot}.

\begin{algorithm}[t]
\caption{Oracle-Based Pairwise Evaluation (Tie Preserved)}
\label{alg:oracle_eval_tie}
\begin{algorithmic}[1]
\Require Dataset $\mathcal{D}=\{(x_i, g_i)\}_{i=1}^{N}$; models $\text{ModelA}, \text{ModelB}$; oracle $J(\cdot,\cdot)$
\Ensure Per-example outcomes $\{\mathcal{E}_i\}_{i=1}^{N}$, win/tie counts $(w_A, w_B, t)$ and rates

\State $w_A \gets 0,\; w_B \gets 0,\; t \gets 0$
\For{$i = 1$ \textbf{to} $N$}
    \State $r_A \gets \text{ModelA}(x_i)$
    \State $r_B \gets \text{ModelB}(x_i)$
    \State $s_A \gets J(r_A, g_i)$ \Comment{Alignment score of A to ground truth}
    \State $s_B \gets J(r_B, g_i)$ \Comment{Alignment score of B to ground truth}
    \If{$s_A > s_B$}
        \State $\mathcal{E}_i \gets r_A$; \quad $w_A \gets w_A + 1$
    \ElsIf{$s_A < s_B$}
        \State $\mathcal{E}_i \gets r_B$; \quad $w_B \gets w_B + 1$
    \Else \Comment{$s_A = s_B$}
        \State $\mathcal{E}_i \gets \textsf{Tie}$; \quad $t \gets t + 1$
    \EndIf
\EndFor
\State \textbf{Compute:} $\text{WinRate}_A \gets \frac{w_A}{N}$,\quad $\text{WinRate}_B \gets \frac{w_B}{N}$,\quad $\text{TieRate} \gets \frac{t}{N}$
\State \textbf{Return} $\{\mathcal{E}_i\}_{i=1}^{N}, (w_A, w_B, t), (\text{WinRate}_A,\text{WinRate}_B,\text{TieRate})$
\end{algorithmic}
\end{algorithm}

\subsection{LLM-based Go Unit Tests Generation}\label{pipeline}

To generate the unit tests from the base model and the fine-tuned model, we use the following three-step workflow to generate the unit test given a source file. This is an important design consideration for our work given that many of the real world Go codebases often exceed the LLM context length. 

\begin{enumerate}[leftmargin=*]
    \item \textbf{Code Chunker}: We have empirically observed when a Go file exceeds 15,360 tokens, prompting an LLM to generate unit tests for the entire file at once becomes inefficient. Despite their advertised large context windows, LLMs are susceptible to the "lost in the middle" issue, which can result in missed functions or logic gaps, ultimately leading to incomplete test coverage ~\citep{liu2024lost}.  
    To mitigate this issue, we implemented a code chunking strategy that processes the file in multiple smaller segments. This approach leverages Abstract Syntax Tree (AST) parsing to construct a dependency tree, ensuring that each chunk preserves logically connected components of the code. Equation \ref{eq:cst} formally describes AST-based chunking, where $T(r)$ represents the AST for code $r$, and $C_i$ denotes the $i^{th}$ code chunk. By systematically partitioning large files into structured and contextually coherent chunks, this method enhances the reliability and accuracy of the generated unit tests while maintaining the integrity of the original code.  
    
    \begin{equation}
    T(r) = \{C_1, C_2, \dots, C_n\}
    \label{eq:cst}
    \end{equation}

    \item \textbf{Unit Test Generation for a Chunk}: 
    For each code chunk \( C_i \), we utilize in-context learning (ICL) for the corresponding unit test generation, denoted as \( \text{ICL}(C_i) \). Through extensive prompt engineering, we have systematically refined our prompts for the test generation accuracy. The prompt template is provided in Appendix, Figure~\ref{fig:prompt_template}.

    \begin{equation}
    UT(C_i) = \{\text{ICL}(C_i)\}
    \end{equation}

    \item \textbf{Compilation of Unit Test File}: Finally, we take the generated unit tests for the chunks and simply append them to give the final unit test file. This can further be enhanced by having an LLM prompt for combining the unit tests. 

    \begin{equation}
    UT(T(r)) = \sum_{i=1}^{n}UT(C_i)
    \end{equation}
\end{enumerate}

\subsection{Setup and Performance Metrics}

\begin{figure}[h!]
    \centering
    \includegraphics[height=6cm]{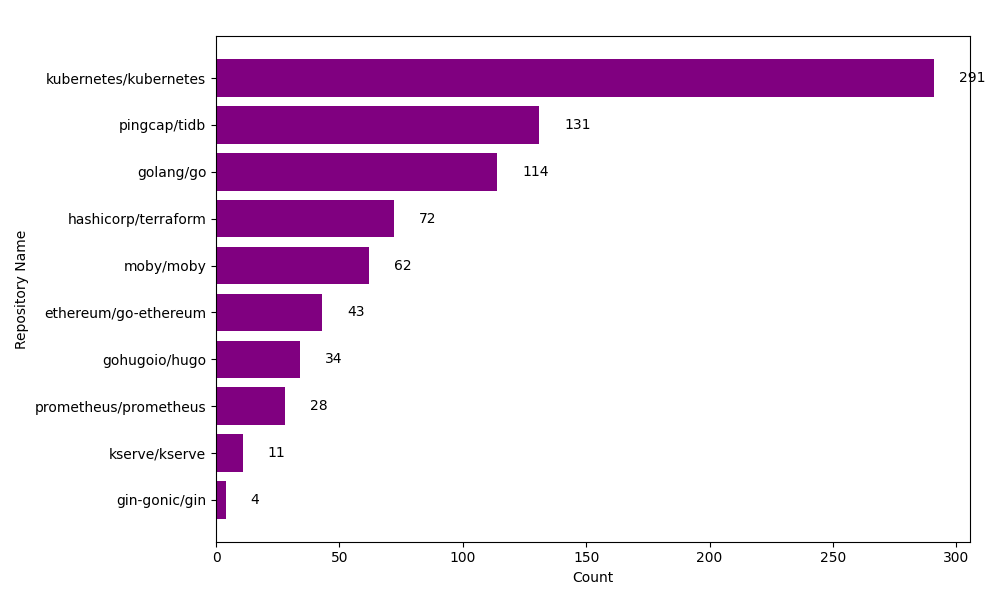}
    \caption{Distribution of validation set samples across repositories in Go-UT-Bench. The dataset contains a total of 790 samples, with Kubernetes having the highest representation (291 samples) and gin-Gonic/gin the least (4 samples), showcasing the varying contributions of different repositories.} 
    \label{fig:validation_set}
\end{figure}

To measure the effectiveness of Go-UT-Bench as a fine-tuning dataset, we conducted fine-tuning experiments on an NVIDIA H100 machine with 80GB VRAM. Without loss of generality, we choose Deepseek-Coder (deepseek-ai/DeepSeek-Coder-V2-Lite-Instruct) ~~\citep{deepseek_hf} and Llama-3.2 (meta-llama/Llama-3.2-3B-Instruct) ~~\citep{llama-3.2} as our test LLMs. Deepseek-Coder is an open-source Mixture-of-Experts (MoE) code language model that comprises 16 billion total parameters—of which 2.4 billion are active during inference—and supports a 128,000 token context length. On the other hand, Llama-3.2 is an instruction-tuned, text-only model. With a context length of 128,000,  Llama-3.2 is optimized for multilingual, on-device dialogue use cases, including agentic retrieval and summarization tasks. The parameter specifications for fine-tuning jobs are mentioned in Appendix: Table \ref{tab:deepseek_fine-tuning} for Deepseek-coder and \ref{tab:llama_fine-tuning} for Llama-3.2. To mitigate overfitting and ensure an unbiased evaluation, we employ a stratified train-test-validation split, allocating 3,684 samples for training and 790 samples each for testing and validation. The dataset distributions across these splits are illustrated in Appendix in Figure \ref{fig:train_set}, Figure \ref{fig:test_set}, and Figure \ref{fig:validation_set}, respectively. These choice strategically covers two LLMs from different model facilities, sizes, and architectures to minimize potential biases.

Following fine-tuning, we evaluate the effectiveness of Go-UT-Bench using the procedure mentioned in Algorithm \ref{alg:oracle_eval_tie}. First, we assess whether the fine-tuned model outperforms the base model and quantify the extent of this improvement. Second, we analyze the distribution of the win-rate across different constituent repositories to identify any significant performance deviations. This assessment is crucial for ensuring consistency and reliability across varying codebases.

\section{Results}
\label{lab:result}

Our results demonstrate that fine-tuning significantly enhances unit test generation. DeepSeek-Coder-V2-Lite-Instruct achieved a win rate of 	81.9\% following Low-Rank Adaptation (LoRA) fine-tuning, a substantial increase from 	14.2\% in its base form. Similarly, Llama-3.2-3B-Instruct exhibited notable improvements, with its win rate rising from 22.0\% (base) to 	76.7\% after full fine-tuning.

\begin{figure}[htbp]
    \centering
    \includegraphics[width=0.95\textwidth]{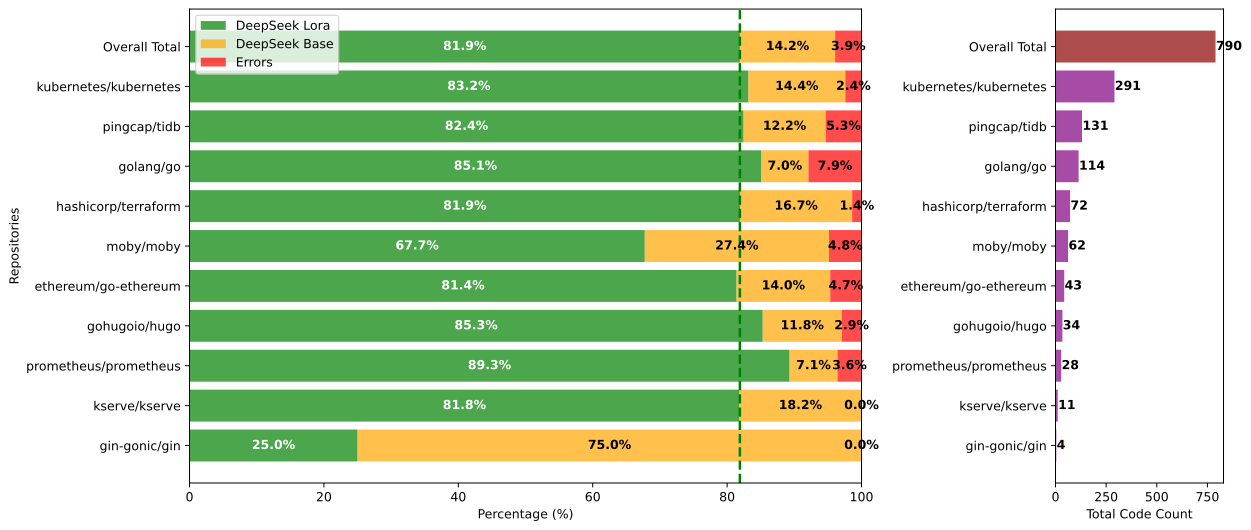}
    \caption{Fine-tuning results for deepseek-ai/DeepSeek-Coder-V2-Lite-Instruct. The left plot compares the win rates between the fine-tuned and the base models across all validation pairs in Go-UT-Bench, while the right plot illustrates the distribution of repositories within the validation dataset.}
    \label{fig:deepseek}
\end{figure}

Figure \ref{fig:deepseek} presents a comparative analysis of the LoRA fine-tuned and base versions of 	deepseek-ai/DeepSeek-Coder-V2-Lite-Instruct. The stacked bars on the left indicate the win rates of both models (fine-tuned vs. base), while the right side exhibits the number of validation pairs per repository. Fine-tuning leads to an 81.9\% win rate over 790 validation pairs, a substantial improvement over the 14.2\% baseline win rate, with an error rate of 	3.9\%. The errors arise from out-of-context issues occurring either during test generation or evaluation. At the repository level, the fine-tuned model surpasses the mean win rate in seven out of ten repositories, including kubernetes/kubernetes, pingcap/tidb, Golang/Go, hashicorp/terraform, GohuGoio/huGo, prometheus/prometheus, and kserve/kserve, demonstrating strong generalization capabilities. However, performance declines for moby/moby (67.7\%) and gin-Gonic/gin (25.0\%), likely due to their lower representation in Go-UT-Bench (62 and 4 validation pairs, respectively). 

\begin{figure}[htbp]
    \centering
    \includegraphics[width=0.95\textwidth]{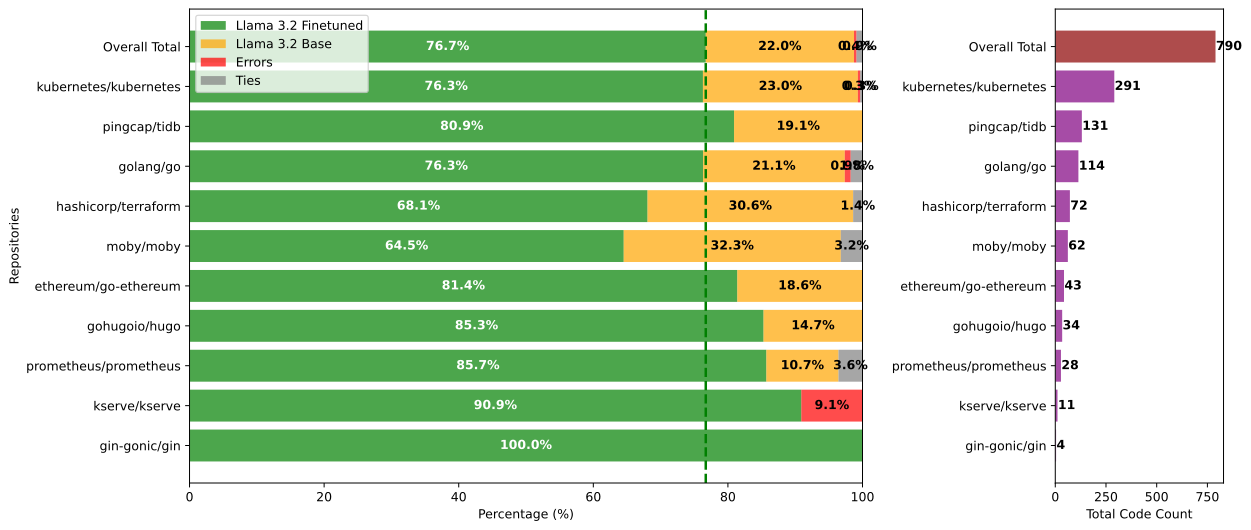}
    \caption{Fine-tuning results for meta-llama/Llama-3.2-3B-Instruct. The left plot compares the win rates between the fine-tuned and the base models across all validation pairs in Go-UT-Bench, while the right plot displays the distribution of repositories within the validation dataset.}
    \label{fig:llama_3-2}
\end{figure}

Figure \ref{fig:llama_3-2} provides an analysis of the fully fine-tuned 	meta-llama/Llama-3.2-3B-Instruct model. Across 790 validation pairs, the fine-tuned model achieves a win rate of 76.7\%, while the base model attains 22.0\%, with an error rate of 1.3\%. At a repository level, the fine-tuned model surpasses the mean win rate of 76.7\% in six out of ten repositories, including {pingcap/tidb}, {ethereum/Go-ethereum}, {GohuGoio/huGo}, {prometheus/prometheus}, {kserve/kserve}, and {gin-Gonic/gin}, further confirming the generalizability of our approach. However, its performance declines in {moby/moby} (64.5\%) and hashicorp/terraform (68.1\%), potentially due to the lower parameter count of Llama-3.2-3B-Instruct.

Taken together, these results underscore the robustness and adaptability of our fine-tuned models across diverse Go repositories. The observed performance improvements demonstrate higher win rates relative to their respective baselines, validating the effectiveness of Go-UT-Bench as a fine-tuning dataset.

\section{Related Work}

A key challenge in adapting large language models (LLMs) to specialized domains lies in the scarcity of high-quality post-training datasets. Supervised fine-tuning (SFT) requires curated instruction–response pairs that align model behavior with downstream tasks. Representative general-purpose resources include FLAN~\citep{longpre2023flancollectiondesigningdata}, Stanford Alpaca~\citep{alpaca}, Dolly~\citep{DatabricksBlog2023DollyV2}, and OpenAssistant~\citep{kopf2023openassistant}, as well as preference-oriented corpora such as Ultrafeedback~\citep{cui2023ultrafeedback} and Anthropic HH-RLHF~\citep{bai2022training}. These datasets primarily broaden the general instruction-following and safety capability surface of LLMs. In contrast, domain-specific resources—such as CodeAlpaca~\citep{codealpaca} for software engineering or MedAlpaca~\citep{medalpaca} for clinical text—remain limited in scale and coverage, underscoring the bottleneck for domain adaptation. 

Recent efforts have begun to address this gap for software engineering. SWE-bench~\citep{jimenez2024swebench} introduces a large-scale benchmark for repository-level issue resolution, while SWT-bench~\citep{mundler2024swtbench} focuses on Python unit test generation. Both benchmarks represent meaningful progress toward realistic software engineering automation evaluation, but they are overwhelmingly Python-centric. Python’s dynamic typing, extensive documentation, and interactive development model make it relatively tractable for dataset construction and LLM adaptation. 

In contrast, Go presents unique challenges that remain underexplored in LLM research. Go is a statically typed and compiled language, where generating correct code requires precise handling of type systems, interfaces, and error propagation. Its concurrency model—built around \emph{goroutines} and \emph{channels}—introduces program behaviors spanning multiple logical hops, making unit test generation significantly more complex than in sequential scripting languages. Moreover, Go repositories frequently contain large, monolithic files, where AST-guided chunking becomes essential for overcoming long-context issues such as the ``lost in the middle'' problem~\citep{liu2024lost}. These structural and semantic characteristics distinguish Go from Python and highlight the absence of domain-specific datasets for fine-tuning LLMs on Go code. n addition, we contribute methodological advances: an AST-guided chunking pipeline to handle long monolithic Go files and a reproducibility-focused dataset design with commit hashes and metadata.

In this work, we contribute Go-UT-Bench, a fine-tuning dataset of \{code, unit test\} pairs from industrial-scale Go projects. Unlike prior Python-centric efforts, Go-UT-Bench explicitly captures concurrency-heavy, statically typed codebases with authentic testing practices, addressing an underexplored yet practically significant programming language.

\section{Discussion and Broader Impact}
\label{lab:conclusion}

This work introduces Go-UT-Bench, a fine-tuning dataset for Go unit test generation with LLMs. Go-UT-Bench consists of 5,264 \{code, unit test\} pairs extracted from ten different open-source repositories, providing sufficient scale and diversity. Our results demonstrate that fine-tuned models significantly outperform their base counterparts in over 75\% of test cases. This highlights the importance of domain-specific adaptation in improving automated test generation.  The findings establish Go-UT-Bench as a valuable dataset for advancing research in AI-driven software engineering and automated testing. Go-UT-Bench includes necessary metadata fields such as commit hash for reproducibility. Beyond serving as a fine-tuning dataset, Go-UT-Bench opens up broader research directions—including cross-language transfer, statically typed unit test generation, and concurrency-aware prompting—and introduces methodological contributions such as an AST-guided chunking pipeline and reproducibility-focused dataset design

Despite its contributions, this study has several limitations.

\begin{itemize}[leftmargin=*]
    \item \textbf{Evaluation Bias:} Our evaluation suffers from several weaknesses: it relies entirely on a single oracle (GPT-4o-mini) to judge outputs, which risks bias due to potential training data overlap with the evaluated models; it lacks triangulation with human judgment or execution-based metrics (e.g., whether generated tests compile or run correctly), leaving the validity of the results open to criticism; it reports only a single measure—win rate—which oversimplifies performance and misses more nuanced breakdowns across repositories, code length, or test complexity; and it does not provide statistical robustness through confidence intervals or significance testing, weakening the reliability of its claims. While LLM-based assessments provide scalability, they may struggle to capture nuanced correctness criteria in complex test cases. Future studies should incorporate human expert evaluations and execution-based evaluation~\citep{xie2024codebenchgen}
    \item \textbf{Data Leakage and Overfitting Risks:} Since Go-UT-Bench is derived from public repositories, there is a risk that some pretrained LLMs have been exposed to portions of this dataset during their training phase. This data leakage could artificially inflate performance metrics, leading to misleading generalization estimates. Future work should implement deduplication strategies and cross-check embeddings against model training corpora to mitigate this issue.
    \item \textbf{Bias in Data Representation:} While Go-UT-Bench encompasses ten diverse repositories, certain domains (e.g., blockchain, API development) are underrepresented, potentially skewing model performance toward highly prevalent domains such as Kubernetes-based infrastructure. Addressing this imbalance is crucial for ensuring fair and comprehensive evaluations.
\end{itemize}

By addressing these limitations and expanding {Go-UT-Bench}, this research aims to contribute as a open fine-tuning dataset for Go unit test generation and help in GenAI-assisted software engineering automation.
While LLM-driven tools can boost perceived productivity, they may inadvertently contribute to developer skill atrophy. For instance, an RCT by Becker et al.~(2025) found that experienced developers using AI coding assistance took approximately 19\% longer to complete tasks—despite believing they were faster—highlighting the risk of relying on AI at the expense of unaided proficiency~\citep{becker2025metr}.

\section{Reproducibility Statement}
\begin{itemize}
    \item Dataset: \url{https://huggingface.co/datasets/Nutanix/GO-UNITTEST-BENCH/tree/main}
\end{itemize}

\pagebreak
\bibliography{neurips_2025}
\bibliographystyle{plainnat}
\newpage

\section{Appendix}

\subsection{Example of Code and Unit test pair from the dataset}
\begin{figure}[h!]
    \centering
    \includegraphics[width=1.25\linewidth]{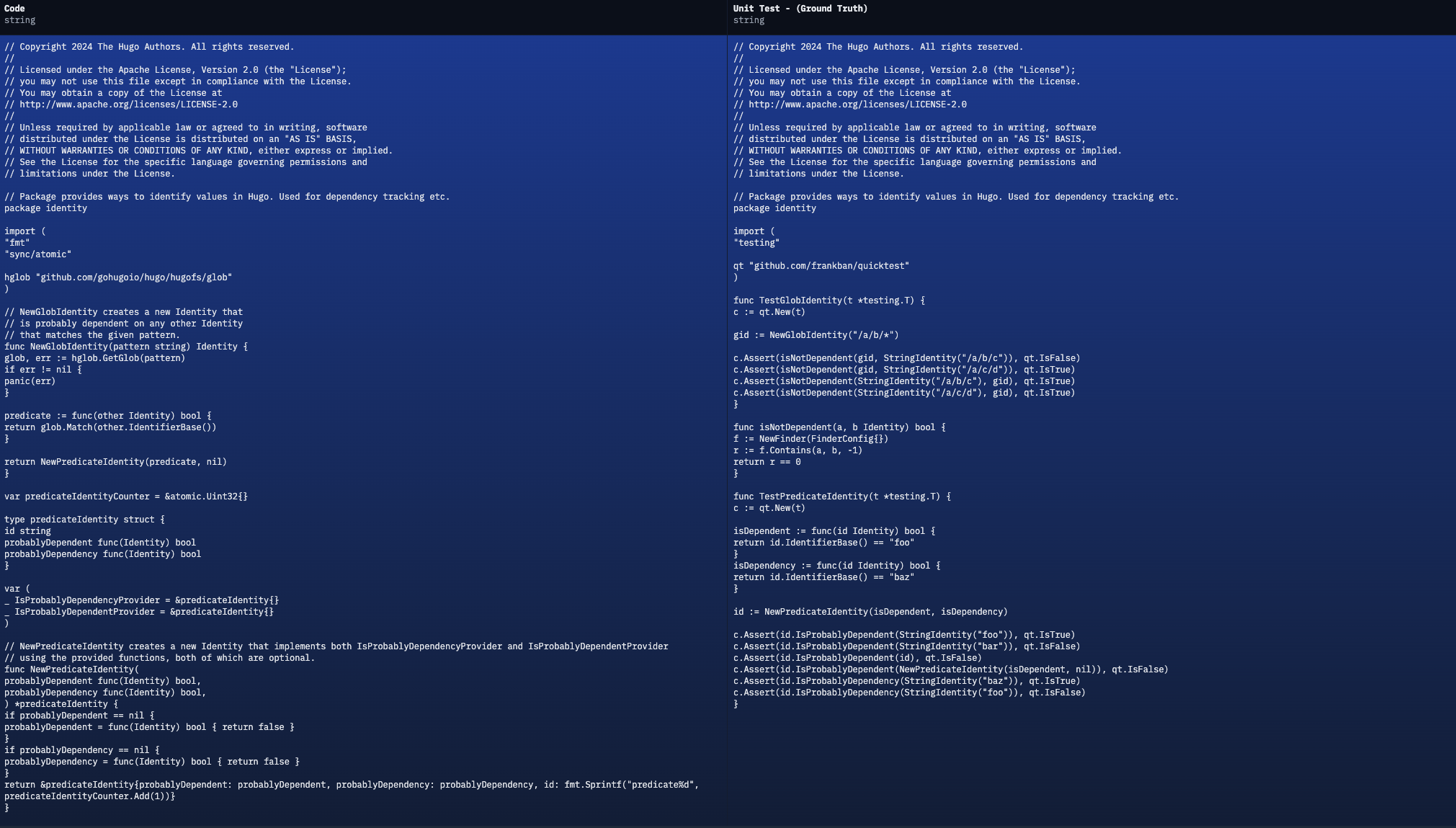} 
    \caption{Example of a code and unit test pair from the dataset}
    \label{fig:code-unit-test-example}
\end{figure}

\pagebreak
\subsection{Unit Test Generation Prompt}

\begin{figure}[h!]
    \centering    \fbox{\includegraphics[width=0.8\textwidth]{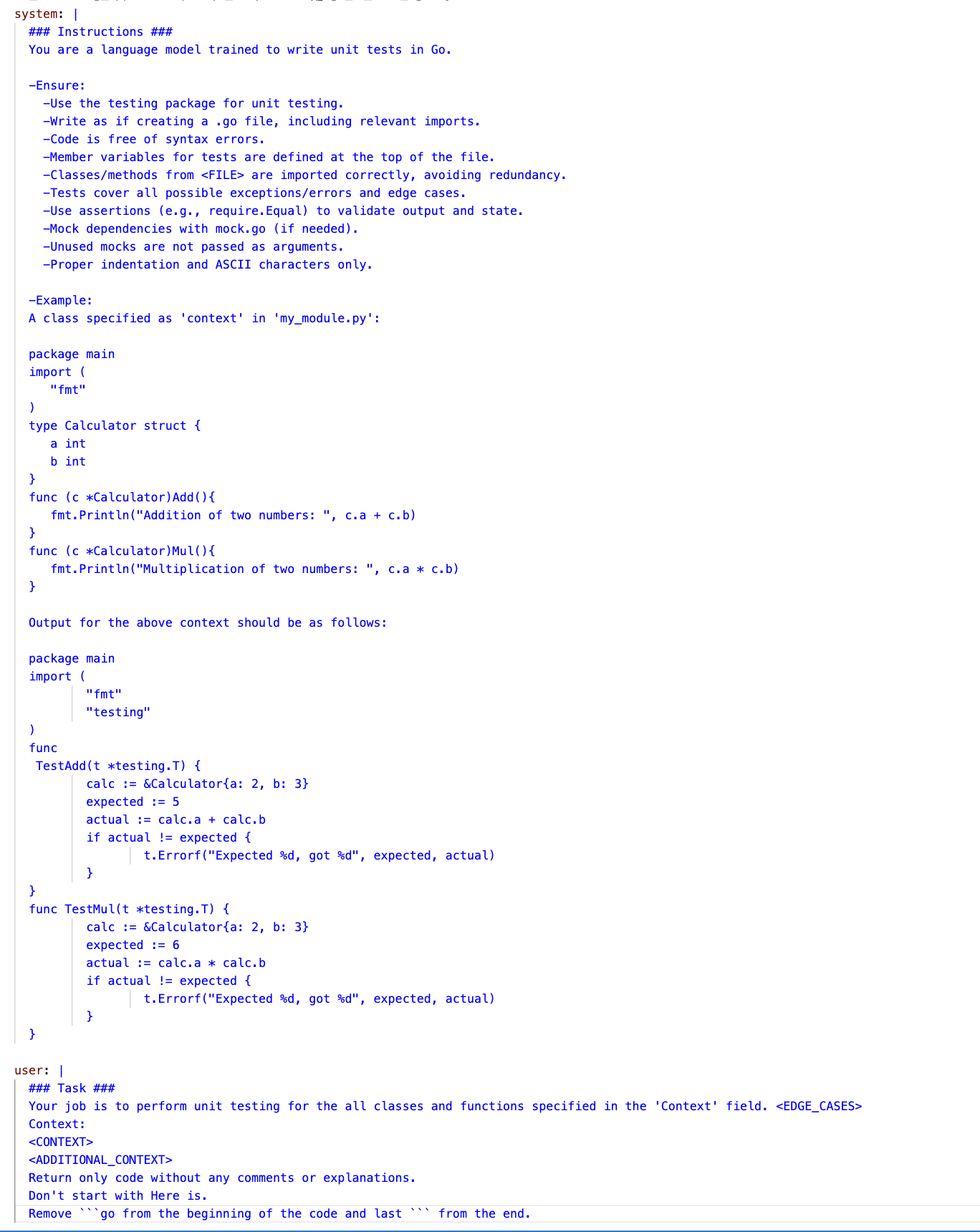}}
    \caption{Prompt template for the unit test generation in Golang. } 
    \label{fig:prompt_template}
\end{figure}

\newpage
\clearpage

\subsection{Evaluation Workflow}
We do comparison between the base-generated unit tests and the LoRA trained DeepSeek-Coder-V2-Lite-Instruct against the Unit test scrapped from the repositories taken as the ground truth. The unit test generated by DeepSeek-Base attempts to capture various edge cases. However, while it includes some relevant edge cases, it fails to achieve full test coverage and also introduces irrelevant edge cases that do not align with the code's functionality. Unit test generated by Deepseek-LoRA however is more concise and covers all the edge cases as well as more closely resembles the Ground Truth. Below is the response from the Judge-Model GPT-4o-mini:

\definecolor{lightergray}{rgb}{0.97, 0.97, 0.97}

\begin{tcolorbox}[colback=lightergray, colframe=black]
To evaluate the responses from Assistant A and Assistant B against the reference answer, we need to consider several factors: correctness, helpfulness, conciseness, and similarity to the ground truth.\par
\vspace{3mm} 
**Correctness:**\par
- **Reference Answer:** The reference answer provides a set of test cases that cover valid local file URLs and invalid non-local hosts. It correctly identifies the expected outputs and error messages.\par
- **Assistant A:** The tests cover valid local hosts and invalid hosts, but it introduces some tests that are not relevant or correct based on the reference (e.g., testing for empty paths and malformed paths). The error messages in some tests do not align with the expected behavior as per the reference.\par
- **Assistant B:** This assistant's tests are more aligned with the reference. It correctly handles both valid and invalid cases, including various scenarios for localhost and non-local hosts. The structure of the tests is clear and follows a logical progression.\par
\vspace{3mm} 
**Helpfulness:**\par
- **Assistant A:** While it attempts to cover various edge cases (like special characters, query parameters, etc.), many of these cases are not necessary based on the original function's intent. This could lead to confusion about what is essential for testing.\par
- **Assistant B:** The tests are straightforward and directly related to the functionality of `convertFileURLPath`. They effectively cover both valid and invalid scenarios without introducing unnecessary complexity.\par
\vspace{3mm} 
**Conciseness:**\par
- **Assistant A:** The response is longer due to additional tests that may not be relevant. This could detract from its overall effectiveness.\par
- **Assistant B:** The response is concise and focused on relevant test cases that directly relate to the function being tested.\par
\vspace{3mm} 
**Similarity with Ground Truth:**\par
- **Assistant A:** While it captures some aspects of the functionality, it diverges significantly by introducing irrelevant test cases that do not align with the reference answer.\par
- **Assistant B:** This response closely mirrors the structure and intent of the reference answer. It includes appropriate test cases for both valid local file URLs and invalid non-local hosts.\par
\vspace{3mm} 
In conclusion, Assistant B's answer is more aligned with the ground truth in terms of correctness, helpfulness, conciseness, and similarity. It effectively captures all necessary scenarios without introducing extraneous tests.\par
\vspace{3mm} 
Final verdict: [[B]]
\end{tcolorbox}

\newpage
\subsection{Parameters used in LoRA-Based Fine-Tuning of deepseek-ai/DeepSeek-Coder-V2-Lite-Instruct}
\begin{table}[h]
    \centering
    \caption{Deepseek-v2 LoRA-Based Fine-Tuning Parameters}
    \label{tab:deepseek_fine-tuning}
    \begin{tabular}{lr}
        \toprule
        \textbf{Parameter} & \textbf{Value} \\
        \midrule
        epochs & 2 \\
        learning\_rate & 5e-05 \\
        batch\_size & 8 \\
        weight\_decay & 5e-02 \\
        grad\_accumulation\_steps & 8 \\
        warmup\_ratio & 0.1 \\
        data\_workers & 8 \\
        max\_seq\_length & 2048 \\
        adam\_epsilon & 1e-08 \\
        lr\_scheduler\_type & "linear" \\
        seed & 42 \\
        logging\_steps & 100 \\
        save\_steps & 1000 \\
        save\_strategy & "steps" \\
        eval\_strategy & "steps" \\
        eval\_steps & 100 \\
        report\_to & "wandb" \\
        gradient\_checkpointing & 'unsloth' \\
        bf16 & True \\
        remove\_unused\_columns & False \\
        packing & True \\
        lora\_ranks & [8] (Looped over this value) \\
        lora\_alphas & [16] (Looped over this value) \\
        lora\_dropout & 0.05 \\
        bias & "none" \\
        task\_type & "CAUSAL\_LM" \\
        target\_modules & ["q\_proj", "k\_proj", "v\_proj", "o\_proj", "gate\_proj", "up\_proj", "down\_proj"] \\
        \bottomrule
    \end{tabular}
\end{table}

\newpage
\clearpage

\FloatBarrier

\subsection{Parameters used in Full Fine-Tuning of meta-llama/Llama-3.2-3B-Instruct}
\begin{table}[h]
    \centering
    \caption{Llama-3.2 Full fine-tuning Parameters}
    \label{tab:llama_fine-tuning}
    \begin{tabular}{lr}
        \toprule
        \textbf{Parameter} & \textbf{Value} \\
        \midrule
        epochs & 2 \\
        learning\_rate & 5e-05 \\
        batch\_size & 8 \\
        weight\_decay & 5e-02 \\
        grad\_accumulation\_steps & 8 \\
        warmup\_ratio & 0.1 \\
        data\_workers & 8 \\
        max\_seq\_length & 2048 \\
        adam\_epsilon & 1e-08 \\
        lr\_scheduler\_type & "linear" \\
        seed & 42 \\
        logging\_steps & 100 \\
        save\_steps & 1000 \\
        eval\_steps & 100 \\
        gradient\_checkpointing & 'unsloth' \\
        bf16 & True \\
        remove\_unused\_columns & False \\
        packing & True \\
        \bottomrule
    \end{tabular}
\end{table}

LoRA (Low-Rank Adaptation) is preferable when computational resources are limited, the model is large (e.g., 7B+ parameters), or the dataset is small to medium-sized, making full fine-tuning impractical. It is particularly effective for domain adaptation, frequent model updates, and avoiding catastrophic forgetting, as it fine-tunes only a small subset of parameters while keeping the rest frozen. In contrast, full fine-tuning is necessary when working with large and diverse datasets, requiring substantial modifications to the model or when optimizing for peak performance in specialized tasks. It is best suited for smaller models ($\leq$3B parameters) or cases where significant re-learning is needed, provided that sufficient GPU resources are available. For scenarios like Go-UT-Bench, LoRA is advantageous for large models such as DeepSeek-Coder-V2-Lite-Instruct, whereas full fine-tuning may yield better results for smaller models like Llama-3.2-3B-Instruct.

\newpage
\clearpage

\subsection{Train-Test-Validation Split}

\begin{figure}[h!]
    \centering
    \includegraphics[height=6cm]{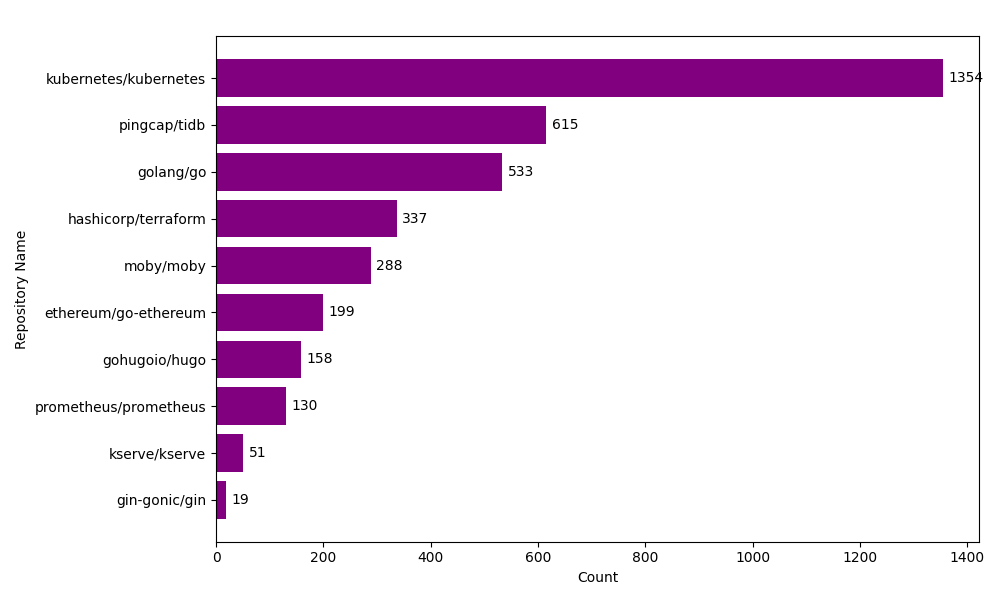}
    \caption{The training dataset consists of 3,684 {code, unit test} pairs extracted from 10 diverse open-source Golang repositories. The figure illustrates the number of pairs contributed by each repository, highlighting the dominance of kubernetes/kubernetes (1,354 pairs), followed by pingcap/tidb (615 pairs) and Golang/Go (533 pairs). Smaller repositories such as gin-Gonic/gin (19 pairs) and kserve/kserve (51 pairs) contribute fewer examples, reflecting the varying availability of unit test coverage across different domains.} 
    \label{fig:train_set}
\end{figure}

\begin{figure}[h!]
    \centering
    \includegraphics[height=6cm]{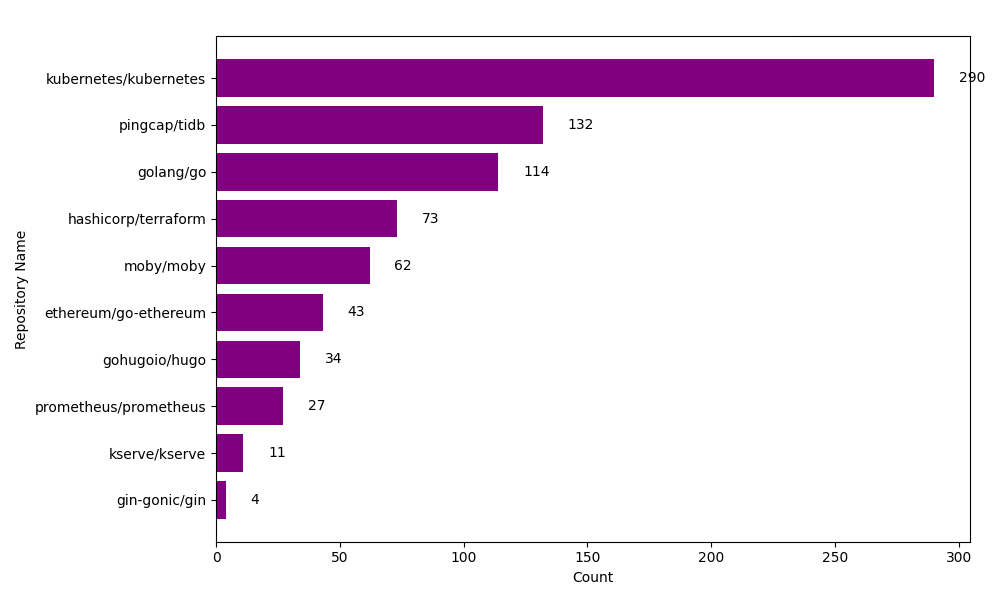}
    \caption{Distribution of 790 test set samples across repositories in Go-UT-Bench. Kubernetes has the highest representation (290 samples), while gin-Gonic/gin has the least (4 samples), reflecting the varying contributions of different repositories.} 
    \label{fig:test_set}
\end{figure}

\newpage

\end{document}